# SleepyWheels: An Ensemble Model for Drowsiness Detection leading to Accident Prevention

Jomin Jose [1], Andrew J [2*], Kumudha Raimond [1], Shweta Vincent [3*]

[1]Department of Computer Science and Engineering, Karunya Institute of Technology and Sciences, Karunya Nagar, Coimbatore, India

[2]Department of Computer Science and Engineering, Manipal Institute of Technology, Manipal Academy of Higher Education, Manipal, 576104, Karnataka, India

[3]Department of Mechatronics, Manipal Institute of Technology, Manipal Academy of Higher Education, Manipal, 576104, Karnataka, India

* Corresponding authors; andrew.j@manipal.edu; shweta.vincent@manipal.edu

## Abstract

Around 40% of accidents related to driving on highways in India occur due to the driver falling asleep behind the steering wheel. Several researches are ongoing to detect driver drowsiness but they suffer with complexity and cost of the models. In this paper, SleepyWheels a revolutionary method that uses a lightweight neural network in conjunction with facial landmark identification is proposed to identify driver fatigue in real time. SleepyWheels is successful in a wide range of test scenarios, including the lack of facial characteristics while covering the eye or mouth, the drivers' varying skin tones, camera placements, and the observational angles. It can works well when emulated to real time systems. SleepyWheels utilized EfficientNetV2 and facial landmark detector for identifying the drowsiness detection. The model is trained on a specially created dataset on driver sleepiness and it achieves an accuracy of 97%. The model is lightweight hence it can be further deployed as a mobile application for various platforms.

## 1. Introduction

As of today, an alarming 40% of highway accidents in India and an exigent 20% in the US occur due to drivers falling asleep behind the steering wheel [1]. Close to 150,000 Indians die annually on the road, of which 60,000 deaths are related to the driver falling asleep behind the wheel. According to the U. S. National Highway Traffic Safety Administration (NHTSA), this issue is responsible for more than 100,000 road accidents, an alarming annual count of 1,500 fatalities and over 71,000 injured victims [2]. These statistics reveal a deep need for timely intervention and careful supervision.

Brake fails, worn-out tires, slippery roads, confusing traffic signs, absence of streetlights and inclined roads on hills are some of the leading causes of road accidents that occur without the direct carelessness of the driver. In contrast to these, forfeiting the use of protective gear, road rage, ignoring the speed limit, distractions, drunk driving, ignoring traffic lights, and driving while sleepy are the leading causes of road accidents that are directly linked to the driver's decision-making and alertness. While road and vehicle maintenance can deal with driver-independent factors and law enforcement can deal with driver-related factors like drunkenness, red light jumping and lack of safety gear, there exists a huge void in the area of keeping a check on how sleepy the driver is, which is overlooked often, but is a leading cause of accidents, nevertheless. This article focuses on monitoring one very important but overlooked factor that compromises the alertness of a driver: sleepiness. Our article outlines the creation and testing of the SleepyWheels app for drowsiness detection in drivers on long-haul travels.

The article is divided into the following sections. Section 2 describes the problem statement, motivation and the uniqueness of our solution. Section 3 presents a brief comparison between state-of-the-art literature and how our system outperforms the systems in the literature. Section 4 gives the detailed architecture and work flow of the SleepyWheels system. Section 5 describes the dataset creation technique for testing and validating the system. Section 6 exhaustively presents the training, deployment and testing scenarios along with the results obtained. Section 7 concludes the article and provides motivation for the scope of future work.

## 2. Motivation & Uniqueness of Proposed Solution

Statistics confirm the common-sense lead conclusion that most drowsiness-related accidents happen between 2 and 5 am [2], [3]. Some people like oil tanker drivers and truck drivers often have no choice but to work at these odd times. While behavioral monitoring [4], ECG [5] and other hands-on forms of supervision require special equipment attached to the driver, which might make driving for long hours uncomfortable. A computer vision-based solution does not impose such requirements. Vehicle or steering monitoring [6] also needs special equipment attached to working vehicle parts while being unable to solve the problem of numerous false positive detections. The main motivation behind choosing computer vision to tackle this problem is its practical advantages such as, no invasive gear that physically interferes with driving and not needing complex equipment except a camera and a speaker.

The past has seen multiple models put to use in unison [7], and isolated models working independently. However, such systems have their shortcomings. Too many parallel machines bring confusion, computational load and lack of decisiveness. Further single deployments of such systems yield unpromising results. The approach of choosing just two models in unison stands out as a well-balanced solution in the load-accuracy trade-off. It is also worth noting that deep learning technology has long been seen as an improver of lifestyle and efficiency of tasks and, as the key to reliable automation. Its scope to save lives has been restricted largely to medical breakthroughs. Hence, this project stands as a testament to prove that technology has the ability to save lives before a victim becomes a patient and reaches the hospital.

A wide assortment of diverse methodologies has been employed in various scientific studies over the past years to solve the problem of drowsiness-induced accidents. The most significant techniques employing physiological signals are ElectroEncephaloGram (EEG) [8], [9], ElectroOculoGram (EOG) [10], [11], ElectroCardioGram (ECG) [5], and ElectroMyoGram (EMG) [12]. These techniques measure electrical activity at various parts of the human body like the brain, the eyes, the heart and the skeletal muscles. The data gathered can be used to derive conclusions about the driver's alertness in real-time. However, these methods require special equipment to be fitted on the driver's body which can be uncomfortable and distracting while driving. Techniques that rely on the state of the vehicle to estimate the driver's alertness keep track of the steering wheel's degree of rotation, accelerator pedal compression and frequency of switching lanes. These approaches assess the driver's driving performance based on real-time vehicle stats and measurement of physiological signals. Deng et al. [13] developed a method that tracks the driver's face by employing facial landmarks to detect lethargy. They checked for signs of sleepiness like yawning, closing of eyes and frequent blinking. Verma et al. [14] used an ensemble of two VGG16s for detecting emotions.

Regions of interest (ROI) from images of the face were fed to the first CNN and facial landmark points were fed to the second. The final decision was a combination of scores from the two models. The architecture used in this solution is outdated but the methodology was a source of inspiration for this work. The authors of [15] tried detecting seven mundane actions performed by drivers like checking mirrors, answering the phone, the activity of setting up in-vehicle video devices, etc. They used a Kinect camera to record the driver's actions. They evaluated 42 features extracted from the video feeds and filtered them based on importance as predicted by random forests. These features were then used to train a feed-forward network that predicts alertness with 80.7% accuracy. Abtahi et al. [16] focused on yawn detection that is unbiased by race, complexion or face wear by developing a sundry dataset that accommodates all edge cases. Their work inspired the diversity of the SleepyWheels dataset. They reported an accuracy of 60%.

The unique solution proposed for the SleepyWheels is to parallelly combine the two best techniques available: facial landmark detection and convolutional neural network classification, in a simple but effective two-threaded ensemble model. The facial landmark detector is Google's BlazeFace, using which aspect ratios of eyes and mouth are calculated to detect sleepiness. The convolutional neural network architecture chosen is the EfficientNetV2B0, which once trained via transfer learning, classifies a frame from the camera's feed as sleepy or alert.

## 3. Literature Survey

Many solutions have been proposed which aimed at tackling the problem of drowsiness detection in drivers. The first class of techniques is based on a comparatively early approach which is based on analyzing the driving patterns. It involves the use of a camera feed to monitor the steering wheel's rotations in order to correlate it with positional deviations from the trajectory predicted by analyzing driving patterns. Krajewski et al. [17] used this relationship between minute steering wheel adjustments to maintain the vehicle's trajectory and driver alertness to achieve 86% accuracy. However, such techniques rely on unreliable variables like the state of the road and vehicle and the driver's skill. The second class of techniques studies the brain and neuromuscular data generated using Electrocardiography (ECG), Electrooculography (EOG), and Electroencephalography (EEG) to judge alertness. These techniques can yield accuracies of over 90% but mandate the attachment of numerous sensors on the body which makes this solution impractical for regular use.

The third class of techniques employs Computer Vision. Periodically snapped stills or video footage from a camera are processed to detect behavioral cues like yawns, frequently dropping one's head, frequent blinks and eyes closed for a long duration. Danisman et al. [18] developed a method to correlate how close the eyelids were to how sleepy the person is. They made the assumption that the number of blinks per minute increases linearly with sleepiness and thus used it to categorize a driver into one of three degrees of sleepiness. Dua et al. [19] modified the Viola-Jones object detection algorithm to measure drowsiness by studying the nature, duration and frequency of yawns. In modern times, techniques that use Convolutional Neural Networks (CNNs) have been very effective for use cases like image classification/ segmentation, object detection/tracking and recognition of emotion. Dwivedi et al. [20] employed shallow CNNs to achieve 78% accuracy in detecting drowsiness in drivers. Park et al. [20] developed a triple-layered architecture employing an AlexNet [21], a VGG-FaceNet

[22] and a FlowImageNet [23] for extracting low-level to complex behavioral features in a step-by-step manner, garnering an accuracy of 73%. Deep learning networks like CNNs detect sleepiness very accurately but typically have high memory and processing requirements which makes their installation on edge devices unideal. However, in recent times, EfficientNets are closing the gap between performance and lightweight models.

Top automobile manufacturers already offer drowsiness detection as a feature in their products. Some examples are Volkswagen's Fatigue Detection System, Renault's Tiredness Detection Warning (TDW), Nissan's Driver Attention Alert (DAA), Jaguar Land Rover's Driver Fatigue Alert and BMW's Active Driving Assistant to name a few. But most of these rely solely on driving patterns formulated based on the driver's interaction with the pedals and steering wheel. The intent of the SleepyWheels research is to prove the efficacy of a simple Computer Vision based ensemble model over the traditional methods used in today's automobile products. Table 1 showcases a comparative analysis of a few systems developed using the idea of computer vision.

Table 1: State-of-the-art Drowsiness Detection Systems

| Ref | Sensor | Features | Algorithm | Metric | Dataset |
|---|---|---|---|---|---|
| Sleepy Wheels | Camera | Video feed of upper body | Ensemble model: landmark detector and CNN | Accuracy: 97% | Self-prepared dataset |
| [24] | Smartphone camera | Frequency and duration of yawns and blinks, pulse, heart beat | Multichannel second-order blind identification | Sensitivity: 94% | Self-prepared dataset |
| [25] | Smartphone | Touch response, PERCLOS, vocal data | Linear Support Vector Machines | Accuracy: 93% | Self-prepared dataset called 'Invedrifac' |
| [26] | EOG, ECG & EEG sensors | Frequency & duration of blinks/closed eyes, heart rate, alpha and beta bands power | Fisher's linear discriminant analysis method | Accuracy: 79.2% | MIT/BIH polysomnographic database |
| [27] | Camera | Facial expression, movements of eyelids, gaze & head | Kalman filtering tracking | AECS: 95% Yawn: 82% PERCLOS: 86% | Self-prepared dataset |
| [28] | Camera | Driver's performance data, movement of the eyes | Support Vector Machines | Accuracy: 81.1% | Self-prepared dataset |
| [29] | PPG, sensor, accelerometer, and gyroscope | Linear acceleration and radian speed of steering wheel, pulse, heart & respiratory rate and variability, stress level | Support Vector Machines | Accuracy: 98.3% | Self-prepared dataset |

## 4. Proposed SleepyWheels System

The proposed solution of the SleepyWheels system parallelly combines outputs of the two of the state-of-the art techniques: facial landmark detection and convolutional neural network classification, in a simple but effective multithreaded ensemble model. The advantage of this approach is two-fold: firstly, the simplicity in design reduces the computation burden on the deployment platform thus opening doors to embedded systems and mobile platforms in contrast to using numerous sensors and multiple ML models in parallel; secondly, this approach provides better accuracy and reduced false positive alarm-triggers that upset the focus of the driver. The facial landmark detector used is BlazeFace [30] from Google's Mediapipe [31]. It provides the information required to calculate aspect ratios of the eyes and the mouth which are used to detect sleepiness. The convolutional neural network model used parallelly is an EfficientNetV2B0 trained via transfer learning, which is trained into a binary classifier that classifies a frame from the camera's feed as sleepy or alert.

### 4.1 Architecture

#### 4.1.1. EfficientNetV2B0 as Binary Classifier

EfficientNet [32] is a convolutional neural network architecture developed by Google. It is the brainchild of a new model scaling methodology whereby a model's depth, width and resolution are scaled uniformly instead of arbitrarily. Version 1 of these networks borrowed MobileNet's inverted bottleneck residual blocks and combined them with the novel squeeze-and-excitation blocks. The EfficientNetV2 [33] family of models has some improvements like the use of fused-MBConv blocks along with regular MBConv blocks. These models are known to achieve higher accuracies with fewer parameters.

EfficientNets also adapt/specialize well when trained on new datasets through transfer learning. Fine-tuned EfficientNets often emerge to be among the state-of-the-art models while using a comparatively lower number of parameters when trained on benchmark datasets like CIFAR-100 (91.7%) and Flowers (98.8%).

In the Sleepywheels app, a custom top is added to a headless EfficientNetV2B0 loaded with weights trained on ImageNet. This network is trained using the techniques of transfer-learning and fine-tuning to develop the binary classification model that decides if a face/upper body shows signs of drowsiness or not. Some important architectural components of this network are explained below.

- Swish Activation

Swish is a multiplication of a linear and a sigmoid activation. It is proven to yield higher accuracies than ReLu in models like InceptionResNet and EfficientNet. The activation function is shown in Eq(1).

$$Swish(x) = x \times Sigmoid(x) \qquad (1)$$

- Inverted Residual/MBConv Block

Originally introduced in the MobileNetV2 [34] architecture, inverted residual blocks start with a widening pointwise convolution, then apply a highly efficient depth-wise convolution which is followed by a compressing pointwise convolution. This approach largely decreases the number of trainable parameters.

- Squeeze and Excitation Block

The Squeeze and Excitation technique grants the network the special ability to treat different channels with different importance. Every channel's importance takes the form of a weight that is treated like a trainable hyperparameter. This new parameter is used to make smarter decisions when judging a multi-channel output, thus leading to higher accuracies.

- Fused-MBConv Block

Performing depth-wise convolutions within MBConv blocks reduces the number of parameters significantly but this does not always translate to improved training speeds because modern accelerators cannot take advantage of such architectural optimizations. When MBConv blocks that are part of the first few layers of an EfficientNet model were swapped with fused MBConv blocks that use regular convolution, experiments reported an improvement in training speed. Hence, researchers recommend the use of a neural architecture search to strike a balance in this efficiency vs training speed trade-off.

### 4.1.2. Facial Landmark Detector

MediaPipe Face Mesh [35] is a landmark detector that can output a stream of estimated locations of important landmarks on the human face like border points of the eyes, nose and mouth in real-time. It can infer the three-dimensional surface of a person's face without the need for a depth sensor.

It uses a combination of a detector model and a landmark generator model. The detector is tasked with providing the landmark generator a crop of the face from the original image. The landmark generator is tasked with predicting the facial surface and assigning new positions to the landmarks.

MediaPipe (Figure 1) has optimized this pipeline so that the landmark generator can use previous facial crops to continue updating the landmark positions in case the face hasn't exited the bounds of the crop. This helps reduce unnecessary use of the detector thus saving power, memory and CPU utilization.

The detector uses a model called BlazeFace. It is a lightweight face detector that performs well on mobile GPUs. It is ideal for real-time face detection since it can run at speeds of over 200 FPS on most mobile devices. It performs feature extraction using a MobileNetV2-based model and anchoring using a Single Shot MultiBox Detector (SSD). It is capable of returning 468 3D facial landmarks in a matter of microseconds.

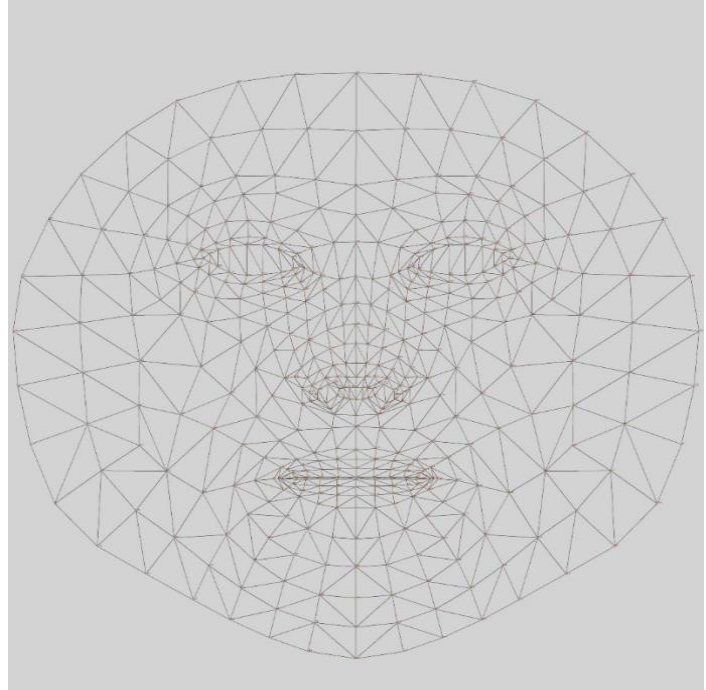

*Figure 1: 468 landmarks returned by MediaPipe's Face Mesh [34]*

The landmarks 30, 29, 28, 243, 22, 24, 463, 258, 259, 359, 254 and 252 are used to calculate EAR (Eye Aspect Ratio) and the landmarks 61, 39, 0, 269, 287, 405, 17, 181 are used to calculate MAR (Mouth Aspect Ratio). Equation 2 calculates the EAR values. Table 2 gives the computed values of the EAR for the left and right landmark and Figure 2 showcases the landmark points for a closed and open eye.

$$EAR = \frac{||p2-p6||+||p3-p5||}{2*||p1-p4||} \qquad (2)$$

Table 2: EAR values of left and right eye landmarks

| Variable | Left Eye Landmark | Right Eye Landmark |
|---|---|---|
| p1 | 30 | 463 |
| p2 | 29 | 258 |
| p3 | 28 | 259 |
| p4 | 243 | 359 |
| p5 | 22 | 254 |
| p6 | 24 | 252 |

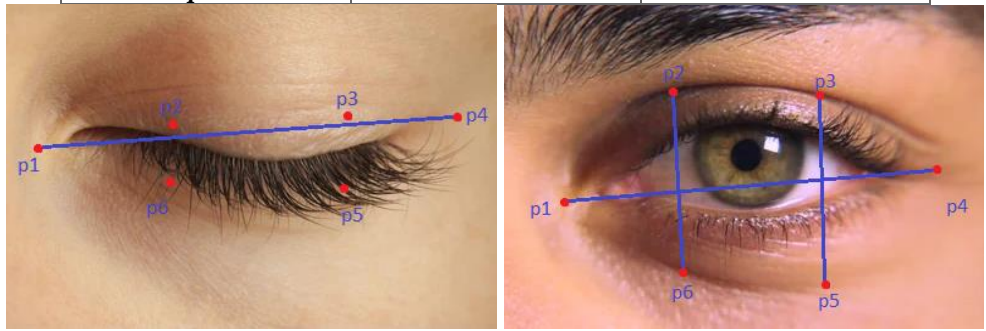

*Figure 2: Measurement of eye aspect ratio (EAR) when closed (left) and when open (right)*

Equation 3 calculates the MAR values. Table 3 gives the computed values of the MAR for the mouth landmark and Figure 3 showcases the landmark points for a closed and open mouth.

$$MAR = \frac{||p2-p8||+||p3-p7||+||p4-p6||}{3*||p1-p5||} \tag{3}$$

Table 3: MAR values of mouth landmarks

| Variable | Mouth Landmark |
|---|---|
| p1 | 61 |
| p2 | 39 |
| p3 | 0 |
| p4 | 269 |
| p5 | 287 |
| p6 | 405 |
| p7 | 17 |
| p8 | 181 |

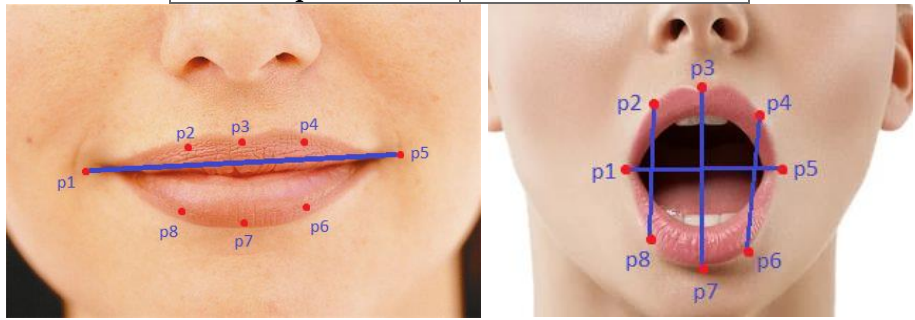

*Figure 3: Measurement of mouth aspect ratio (MAR) when closed (left) and when open (right)*

## 4.2 Workflow

The overall workflow of the SleepyWheels system is shown in Figure 4. One video frame of the driver is captured by the camera (smartphone/webcam/independent camera) and sent to OpenCV. Two threads are created and deployed to run parallelly: one for the neural network and one for the facial landmark detector. In the first thread, a series of functions transform the image into the right size for the neural network, and it is fed into the neural network which returns a classified value between 0 and 1. This value represents the probability of how sleepy the driver could be. In the second thread, the image is processed by Mediapipe's Face Mesh, which returns a set of landmarks that are used to calculate eye and mouth aspect ratios. The aspect ratio of the mouth is used to determine if the driver is yawning. If so, the database is triggered to register the yawn. The eye's bounding points are used to assess how sleepy the driver is, by calculating aspect ratios and comparing with predefined threshold values. The results of the two threads are combined to decide whether to trigger the alarm or not. If the alarm is triggered, the database is updated with a timestamp. Finally, at a later point of time, the driver can access the web dashboard to view how sleepy he or she was while driving.

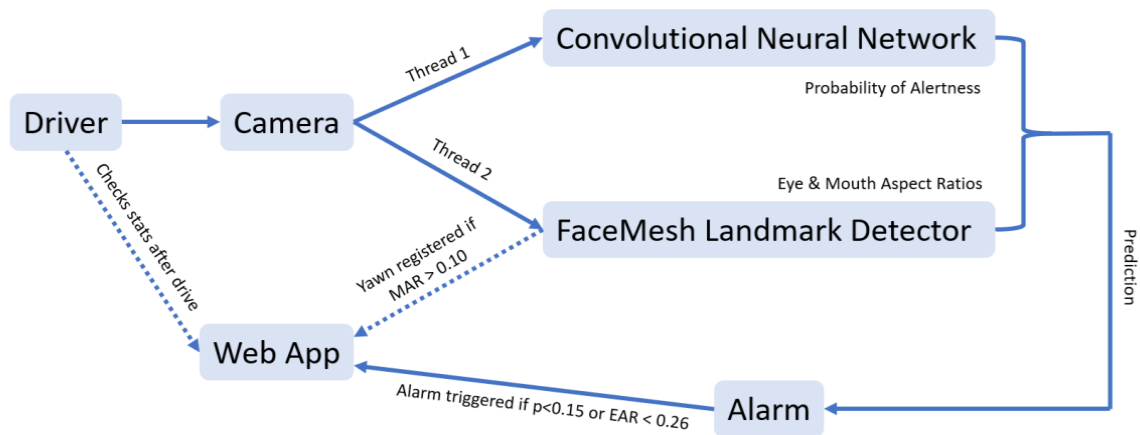

*Figure 4: Workflow of SleepyWheels system*

## 5. Dataset Creation

The image database used for transfer learning with the EfficientNetV2 convolutional neural network was custom-made by crawling Google Images. Ever since Google made changes to its Image Search API to prevent mass web crawling for images, image crawling on the world's biggest image database has not been easy. Nevertheless, an image crawling program by Jayden Oh Yikong hosted on Github, called Google-Image-Scraper was used to download images of sleepy or alert people. A python script in this scraper takes a list of search keywords such as:

- drowsy/sleepy boy/girl/woman/man/driver/portrait (with glasses/mask)
- yawning boy/girl/woman/man (with glasses/mask)
- clumsy/tired/slumber/doze off
- alert/awake boy/girl/woman/man/driver/portrait (with glasses/mask)
- portrait of a boy/girl/woman/man (with glasses/mask)

and the number of images to crawl as input arguments. It uses Selenium and Google Chrome browser and driver to navigate to the Google Images webpage, type in the keywords turn by turn, downloads pictures in PNG and JPEG formats and scrolls down to load more images. The downloads were a mixture of both relevant and irrelevant images, and hence a process of quickly going through each batch of downloads and manually deleting irrelevant images had to be done to ensure the purity of the data being supplied to the neural network for training. The sample data set is illustrated in Figure 5. The details of the number of images used for training and validation are is shown in Table 4.

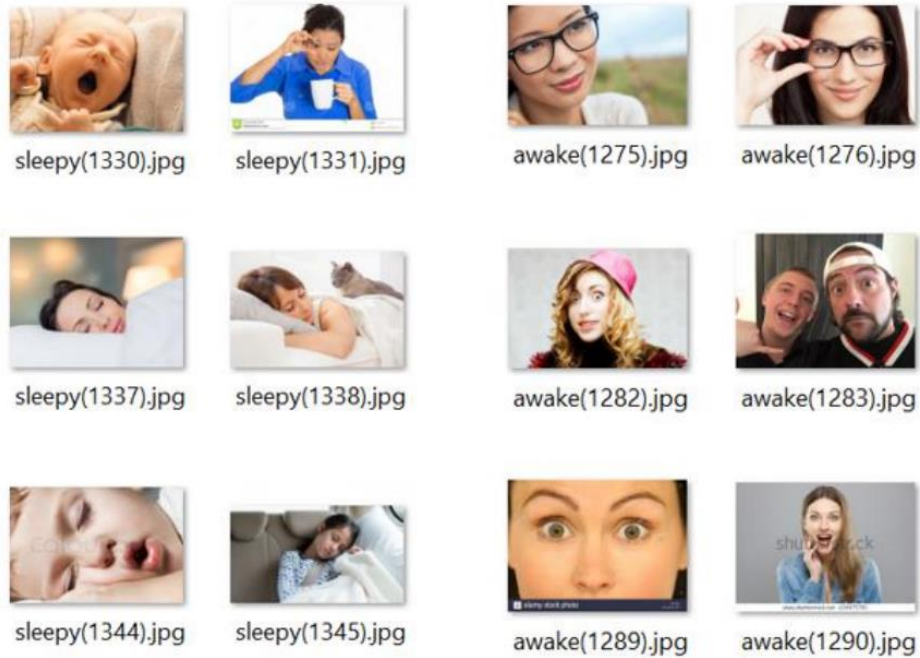

*Figure 5: Examples of (Left) Sleepy and (Right) Awake images in the dataset (Adapted from Google Image Scraper)*

About 16,000 images were downloaded, out of which 10,076 were chosen after a careful manual filtering process. The test-train split was performed manually, and two folders were created named train and validation, each housing its own two folders for sleepy and alert images. The train folder contains 9068 images, 4534 belonging to each class, while the validation folder contains 1008 images, 504 of each class. The Google Drive link to the dataset is provided in [36].

*Table 4: Dataset Summary*

| Type | Number |
|---|---|
| Train - sleepy | 4534 |
| Train - awake | 4534 |
| Validation - sleepy | 504 |
| Validation - awake | 504 |
| **Total** | **10076** |

## 6. Training, Deployment and Testing

This section of the article provides an insight into the training procedure followed to implement the SleepyWheels system.

### 6.1 Initial training

The machine learning frameworks of Tensorflow and Keras have been used in this project. Before training, the bottom layers of the EfficientNetV2 B0 network are locked with the pretrained weights from ImageNet challenge, and a max-pooling layer, a flattening layer, a dropout layer, a fully connected layer of 1024 neurons that use ReLU activation and an output layer consisting of one neuron with sigmoid activation has been added to the top. At this point, the number of trainable parameters was have been 64.2 million. Figure 6 gives a snapshot of the model summary just before performing the initial transfer learning.

Table 5: Hyperparameter Tuning Details

| Stat/Hyperparameter | Original | Before Transfer Learning | Before Fine Tuning |
|---|---|---|---|
| Total parameters | 7.1 million | 70.15 million | 70.15 million |
| Trainable parameters | 7.1 million | 64.22 million | 70.08 million |
| Non-trainable parameters | 0 | 5.91 million | 61 thousand |
| ReduceLROnPlateau | warmed up from 0 to 0.256 then decayed by 0.97 every 2.4 epochs | factor=.01, patience=3, min_lr=1e-4 | factor=.01, patience=3, min_lr=1e-5 |
| Initial learning rate | | 3e-3 | 3e-4 |
| Optimizer | Adam | Adam | Nadam |
| Momentum | 0.9 | 0.9 | |
| Momentum Decay | 0.9999 | 0.999 | |
| Batch size | 4096 | 64 | |
| Epochs | 350 | 50 | |
| Neurons in penultimate layer | - | 1024 | |
| Activation at penultimate layer | - | ReLU | |
| Dropout at penultimate layer | - | 0.7 | |
| Image Augmentation | RandAugment [37], Mixup [38] | o rescale=1./255,<br>o rotation_range=180,<br>o width_shift_range=0.1,<br>o height_shift_range=0.1,<br>o shear_range=0.1,<br>o zoom_range=[0.9, 1.5],<br>o brightness_range=[0.5,1.1],<br>o horizontal_flip=True,<br>o vertical_flip=True,<br>o fill_mode='nearest' | |

*Figure 6: Model summary just before initial transfer-learning*

The CNN's bottom layers were left frozen to avoid changing the core convolutional layers that were responsible to detect features in the images. Keeping them frozen at earlier stages helps to train the top classification layers better. An Nvidia RTX 3060 Ti GPU connected to a computer running Ubuntu was used for remote training via SSH. This model was trained on the custom 10,000 image SleepyWheels dataset, for 32 epochs with image size 224x224, learning rate of 0.00002, RMSProp optimizer, using binary_accuracy as metric, binary cross entropy as loss function, batch size 64 and a dropout rate of 0.6 on the flattened layer. Tensorboard was used to monitor performance metrics. At this stage, the model achieved training loss and accuracy of 0.5795 and 85% respectively and a test loss and accuracy of 0.6383 and 84% respectively. The accuracy and loss plots are illustrated in Figures 7 and 8.

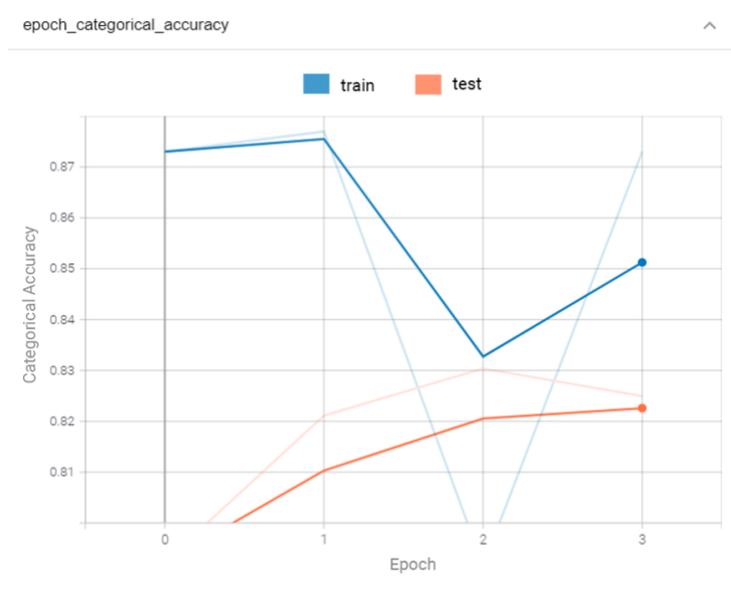

*Figure 7: Accuracy plot during transfer learning*

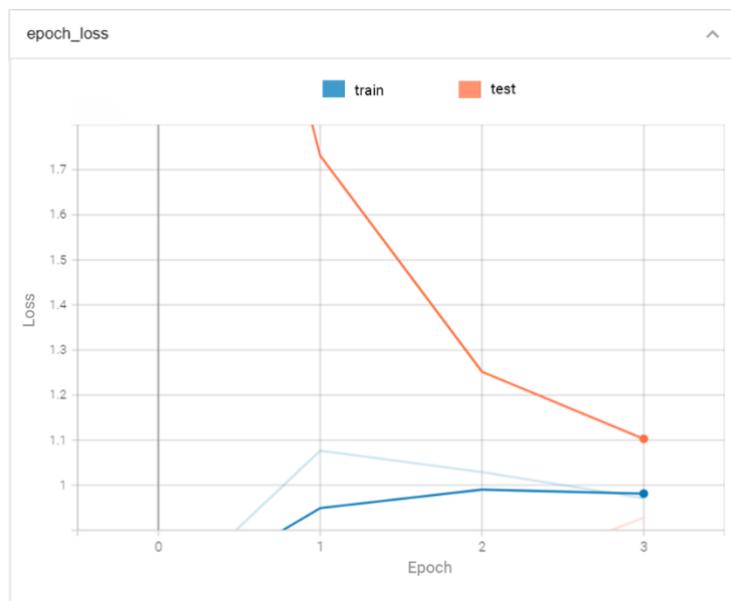

*Figure 8: Loss plot during initial transfer learning*

## 6.2 Finetuning of Performance

The next step was to fine tune the performance of the network in which all the layers of the EfficientNetV2B0 model were set to trainable. This unlocked about 5 million previously untrainable parameters so as to try and achieve a bit more specialization to the current classification problem. The model parameters just before fine tuning are presented in table 5.

The model was trained using the checkpoint callback to make sure that the best version of the model during training was saved to Google Drive. The model was trained on the same train and validation images, with rescaling, rotation, width/height shift, shear, zoom, brightness, horizontal/vertical flip methods used for data augmentation, binary cross entropy as loss function, accuracy as metric, learning rate of 0.0001 with Adam optimizer and batch size of 64 for 50 epochs. Tensorboard was used to monitor performance metrics. Final training and testing accuracies were 97% and 96.92%. The model achieved about 12% more accuracy through fine-tuning the details of which are illustrated in Figures 10, 11 and 12. The confusion matrix obtained has been tabulated in Table 5. Further, Figure 13 shows a sample of some of the misclassified images.

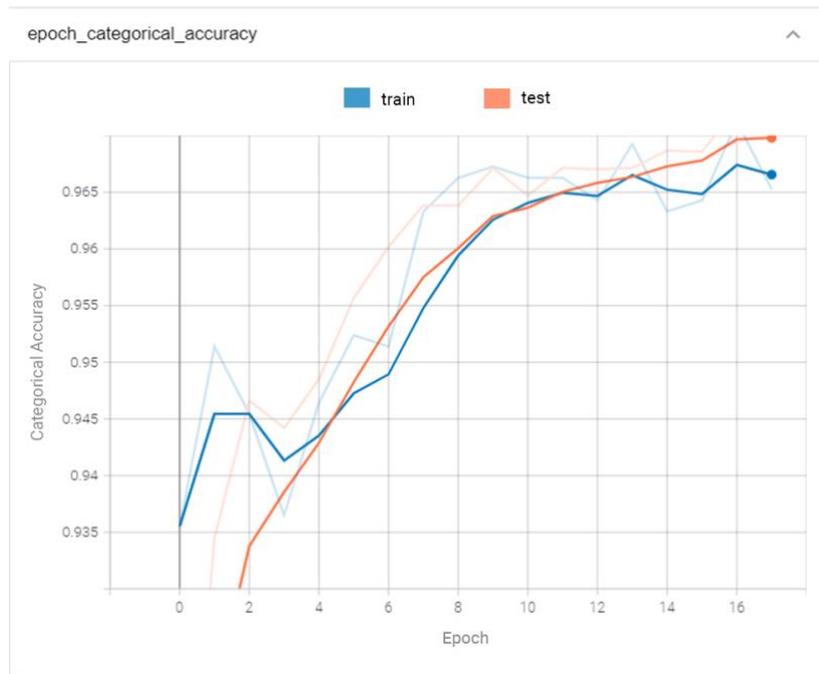

*Figure 10: Accuracy plot during fine-tuning*

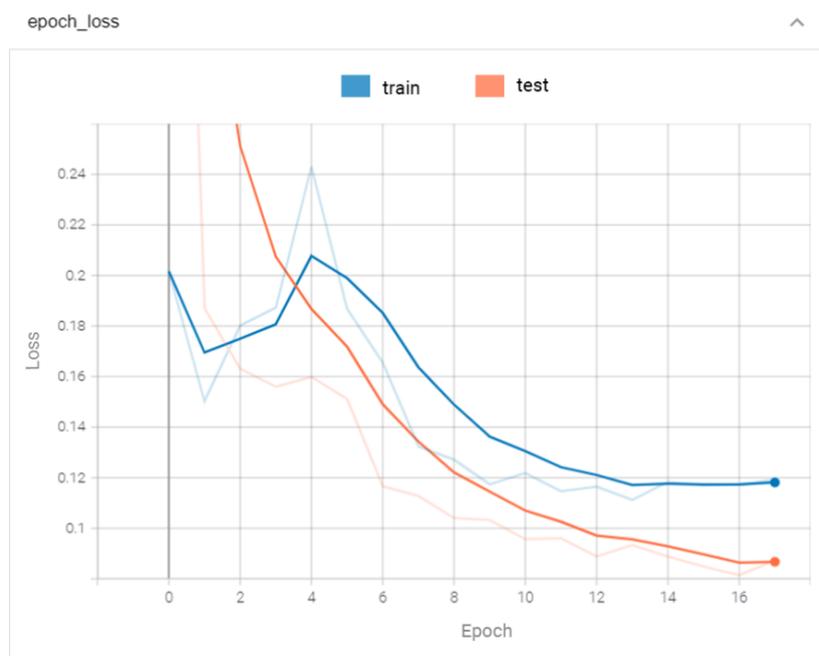

*Figure 11: Loss plot during fine-tuning*

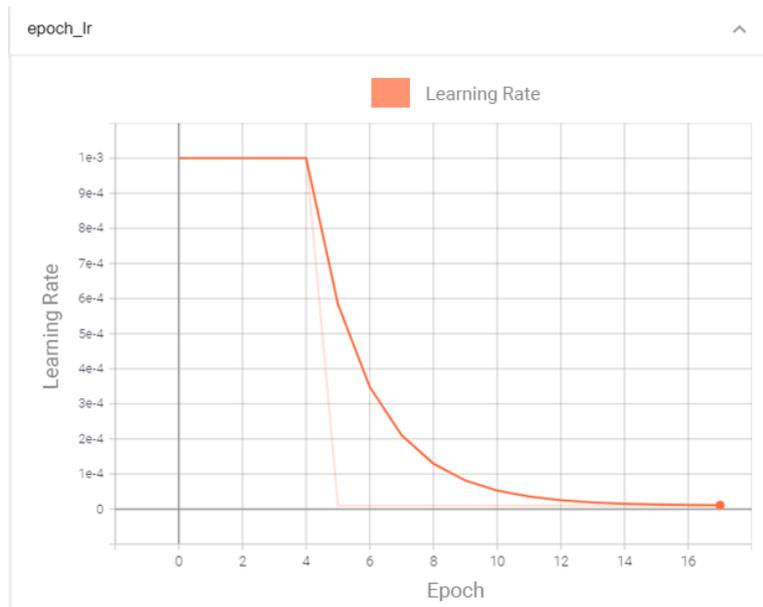

*Figure 12: Learning rate plot during fine-tuning*

*Table 5: Confusion Matrix*

|  | **Awake** | **Sleepy** |
|---|---|---|
| **Awake** | 498 | 15 |
| **Sleepy** | 18 | 481 |

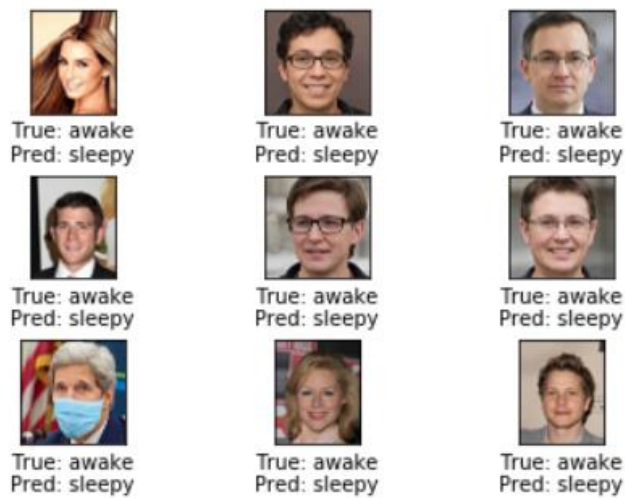

*Figure 13: Examples of images that were misclassified by the model*

The next step was to perform the deployment of the SleepyWheels framework which is described in the following section.

## 6.3 Deployment

The deployment of the SleepyWheels framework was performed in two steps which are described below.

### 6.3.1. Frame-processing loop

The first step is to read a frame from the camera. Next, this frame is resized and fed to the convolutional network on a separate thread. On the main thread, the frame is resized to a smaller width while maintaining the aspect ratio and converted to grayscale. This is fed to the facial landmark detector, which returns a list of rectangles where faces were found. Each rectangle is processed and landmarks returned are stored in a NumPy array. These landmarks are used to calculate aspect ratios of the eyes and mouth using the utility functions mentioned above. During debugging, landmarks are used to plot the contours around the eyes and mouth. The mouth's aspect ratio is used to detect yawns. The average of left and right eyes' aspect ratios is used to decide if the eyes are closed for long enough to decide that the person is sleepy. This result is coupled with the probability returned by the neural network, and a decision is made whether to increment the frame counter, which counts the number of frames in which the person was detected to be sleepy. When this counter exceeds the threshold (currently set as 60), the alarm function is called on a separate thread.

### 6.3.2. Dashboard's backend

Keeping an XAMPP server running, a PHP file called 'index.php' is written into htdocs folder of xampp. This page contains a basic UI which displays the yawn and alarm total counts and two tables for the respective entries accumulated. On PHPMyAdmin, a basic MySQL database called 'sleepywheels' is created, and two tables, yawns and alarms are created in it, each containing two columns: one column to represent the string 'yawn' or 'alarm' and one timestamp column. Two additional PHP pages are created, to which yawn and alarm data are posted by the frame processing loop. Once debugging is done, the database and the PHP files are hosted on 000webhost.

## 6.4 Live Testing

This section of the article showcases the live testing scenario and results obtained. Figures 14 illustrates the cases when the driver has been sleeping leading to the alarm being triggered and when the driver is awake. Figure 15 showcases how the SleepyWheels system is able to detect sleepiness even if facial features are not detected due to a covered or turned face.

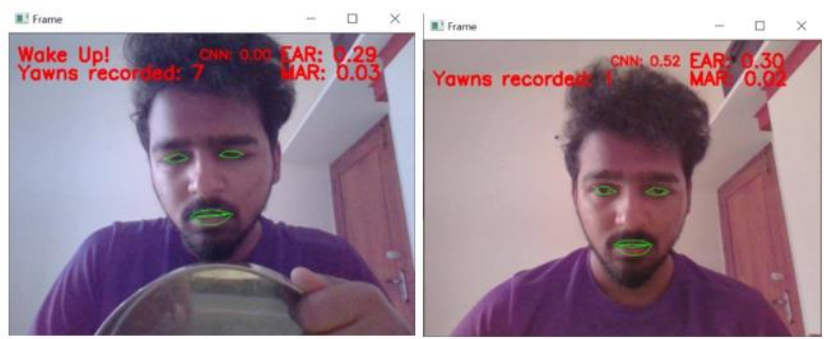

*Figure 14: Alarm is triggered (left) | Driver is alert (right)*

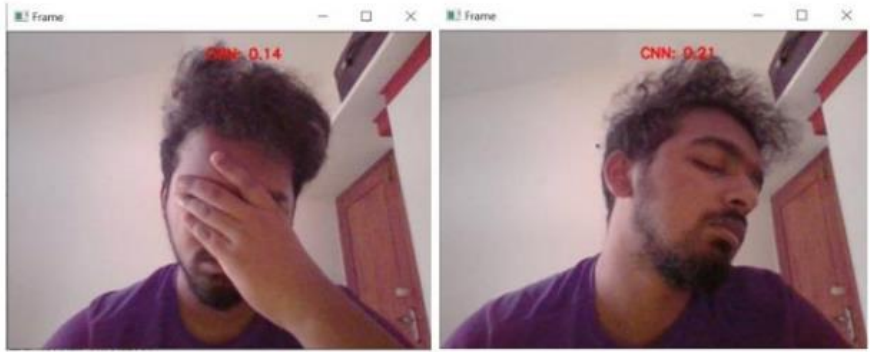

*Figure 15: Ensemble model detects sleepiness in spite of missing facial features*

During two one-hour tests, SleepyWheels correctly triggered the alarm 95% of the time and triggered false positive alarms just 1% of the time. Hence it proves to be a system worth deploying in real-life scenarios. As anticipated, the landmark detector works with the CNN to better predict whether the driver is sleepy or not thus minimizing false alarms which would be counterproductive by being a distraction from driving. Also, in cases where the eyes or the mouth cannot be detected by the landmark detector, the CNN steps in and saves the model from wrong results by providing a response, in contrast to models that use the detector alone which cannot function reliably without the availability of facial features in the frame. The backend database updates in real-time and the webapp displays the list of alarm and yawn triggers to the driver. Figure 16 illustrates the SleepyWheels web dashboard with details of the activity recorded.

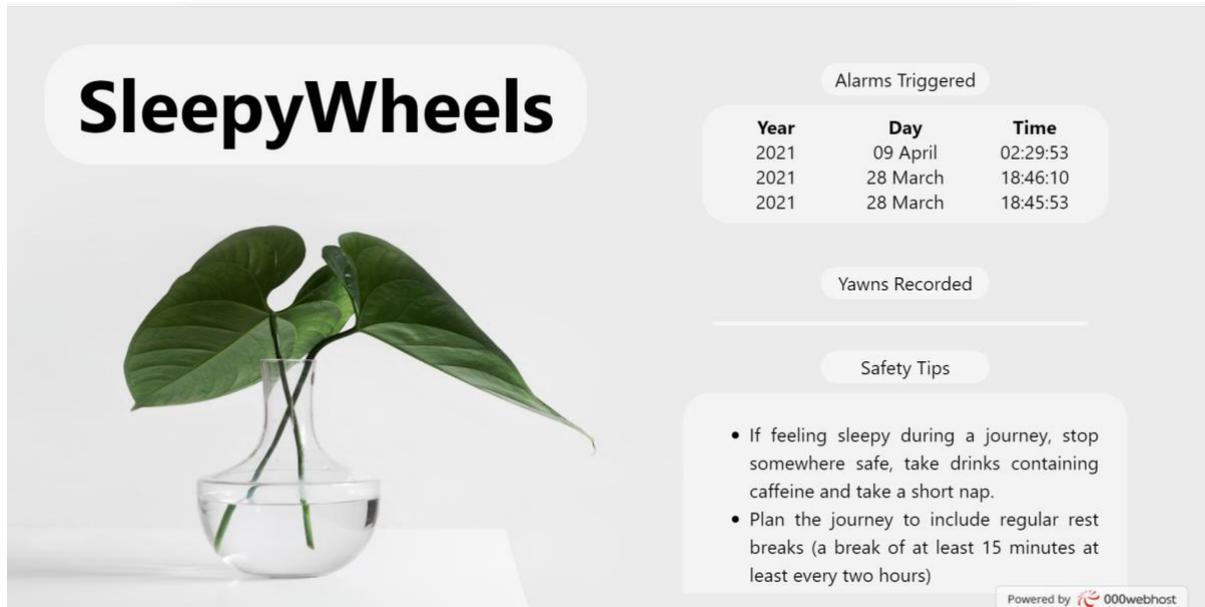

*Figure 16: SleepyWheels Web Dashboard*

## 7. Conclusion

SleepyWheels is a novel approach toward drowsiness detection using a lightweight convolutional network in parallel with facial landmark detection, to achieve real-time driver drowsiness detection. It achieves an accuracy of 97% on the custom-made SleepyWheels dataset. It has proved to be effective in a variety of test cases, such as absence of facial features while covering eye or mouth, varying skin complexion of drivers, varied positions of the camera and varying angles of observation. It has the potential to be deployed successfully in real cases. It has a minimal computational load on the system, which makes it suitable to be shipped to mobile platforms in the form of Android and iOS apps. It has the potential to be further improved through code optimizations in the frame processing loop. Overall, it is a successful attempt at combining two of the best-known ways to tackle the problem to produce better results.

**Declarations**

**Competing interests**
We declare no conflict or competing interests.

**Authors' contributions**
J.J. and A.J. Conception and design of the work, J.J., A.J., and K.R. Data collection, J.J., A.J. Data analysis and interpretation, A.J. and K.R. supervised the experiments, J.J., A.J., and S.V. drafting the article, S.V. and A.J Critical revision of the article, all authors have approved the final version of the article.

**Funding**
There is no funding received for this research work.

**Availability of data and materials**
The data and code used for this research are available in the following GitHub link.

https://github.com/jominjose14/SleepyWheels

https://drive.google.com/drive/folders/1bhrgY8RcUFuD675oxcSLJkmtxY3Wxfg9 (accessed Aug. 08, 2022).